\def\year{2022}\relax
\documentclass[letterpaper]{article} 
\usepackage{aaai22}  
\usepackage{times}  
\usepackage{helvet}  
\usepackage{courier}  
\usepackage[hyphens]{url}  
\usepackage{graphicx} 
\usepackage{natbib}  
\usepackage{caption} 
\usepackage{booktabs}
\DeclareCaptionStyle{ruled}{labelfont=normalfont,labelsep=colon,strut=off} 
\usepackage{algorithm}
\usepackage{algorithmic}
\usepackage{multirow}
\usepackage{xcolor}

\usepackage{newfloat}
\usepackage{listings}

\floatstyle{ruled}
\newfloat{listing}{tb}{lst}{}
\floatname{listing}{Listing}
\title{Best of Both Worlds:\\A Hybrid Approach for Multi-Hop Explanation with Declarative Facts}
\author{
    Shane Storks\textsuperscript{\rm 1}\thanks{Work completed during an internship with Amazon Alexa AI.},
    Qiaozi Gao\textsuperscript{\rm 2},
    Aishwarya Reganti\textsuperscript{\rm 2},
    Govind Thattai\textsuperscript{\rm 2}
}
\affiliations{
    \textsuperscript{\rm 1}University of Michigan\\
    \textsuperscript{\rm 2}Amazon Alexa AI\\
    sstorks@umich.edu, \{qzgao, areganti, thattg\}@amazon.com
}

\begin{document}

\maketitle

\begin{abstract}
Language-enabled AI systems can answer complex, multi-hop questions to high accuracy, but supporting answers with evidence is a more challenging task which is important for the transparency and trustworthiness to users. Prior work in this area typically makes a trade-off between efficiency and accuracy; state-of-the-art deep neural network systems are too cumbersome to be useful in large-scale applications, while the fastest systems lack reliability. In this work, we integrate fast syntactic methods with powerful semantic methods for multi-hop explanation generation based on declarative facts. Our best system, which learns a lightweight operation to simulate multi-hop reasoning over pieces of evidence and fine-tunes language models to re-rank generated explanation chains, outperforms a purely syntactic baseline from prior work by up to 7\% in gold explanation retrieval rate.
\end{abstract}

\section{Introduction}

Efficient and natural human communication relies on implicit shared knowledge and underlying reasoning processes. Despite rapid progress in language-enabled AI agents for tasks like question answering and more, state-of-the-art systems still struggle to explain their decisions in natural language. To improve their interpretability and robustness, a number of multi-hop explanation generation and identification benchmarks based on large, unstructured corpora of facts have been created~\cite{mihaylov2018can,khot2020qasc,jhamtani-clark-2020-learning}. However, when generating explanation chains, powerful deep neural networks can be too cumbersome to use in large-scale applications, while the fastest systems lack reliability, as they depend on syntactic features and ignore semantic relations between concepts~\cite{banerjee2020knowledge,jhamtani-clark-2020-learning}. In this work, we present novel approaches to integrate efficient syntactic retrieval methods with flexible semantic modeling methods for multi-hop explanation. 
Our methods simulate a multi-hop reasoning process from the retrieval and synthesis of evidence to re-ranking candidate explanations.



\section{Related Work}


Recent work has focused on different aspects of multi-hop reasoning for question answering and related natural language understanding tasks. One line of work has incorporated highly structured knowledge graphs into language understanding by 
combining graphical methods with language models~\cite{lin-etal-2019-kagnet,ji-etal-2020-language,yasunaga-etal-2021-qa},
augmenting language model inputs with relational knowledge~\cite{zhang-etal-2019-ernie,chen-etal-2020-improving,xu-etal-2021-fusing},
and applying language models to relational knowledge to infer multi-hop reasoning paths through knowledge graphs \cite{wang-etal-2020-connecting}. Others have further explored training language models with semi-structured relational knowledge \cite{sap2019atomic,bosselut-etal-2019-comet,mostafazadeh-etal-2020-glucose,Hwang2020COMETATOMIC}, i.e., where nodes are natural language sentences rather than canonicalized concepts, to later use for generating multi-hop explanations in natural language \cite{shwartz-etal-2020-unsupervised,Bosselut2019DynamicKG}. 

For generating multi-hop explanations from entirely unstructured corpora, other work has explored using multi-step syntactic information retrieval methods \cite{jhamtani-clark-2020-learning}, and modeling such corpora as knowledge graphs with relations induced by shared mentions of concepts between documents \cite{dhingra2020differentiable,lin-etal-2021-differentiable}. While the former approach lacks the ability to capture semantic relationships between evidence sentences, the latter demands high time and space complexity both in generating a graph from corpora of millions of facts, and in everyday uses of adding or removing facts from the corpus. 
More recent work has used pre-trained word embeddings to add some lightweight semantic representation to syntactic evidence retrieval \cite{yadav-etal-2021-want}.
Unlike these approaches, we present a flexible and relatively lightweight pipeline to apply both syntactic and learned, contextualized semantic approaches in multi-hop explanation generation, including evidence retrieval, multi-hop reasoning over evidence, and re-ranking candidate explanations.




\section{Problem Statement}
In the research community, two types of explanation have been studied: introspective explanation and justification explanation~\cite{biran2017explanation}. The former explicates how a decision is made, and the latter gathers evidence to support a decision. In this study, we focus on the task of justification explanation. Specifically, we explore the problem of generating multi-hop explanations to support the answer to a natural language question, where the explanation chain is generated from an unstructured corpus of declarative facts. Unstructured natural language corpora are suitable knowledge resources for human-AI interaction, as humans can easily support reasoning by providing their own commonsense knowledge in short, natural language statements. This carefully restricted problem of explanation generation consists of two key challenges. First, we must solve the \textit{retrieval} task to gather candidate supporting evidence from the corpus. Second, we need to invoke a \textit{multi-hop reasoning} process to connect pieces of evidence to form the most valid explanation to justify the answer to the question.






\subsection{Datasets}
To explore this problem, we consider two datasets. First, the Question Answering via Sentence Composition (QASC) dataset provides about 10,000 multiple-choice science questions \cite{khot2020qasc}. QASC is a challenging problem, as each question requires composing two facts from a corpus of about 17 million declarative facts to connect the question and its correct answer. 
For example, given the question ``\textit{Differential heating of air} can be harnessed for what?'' and correct answer ``\textit{electricity production},'' the answer can be explained by composing the facts ``\textit{Differential heating of air} produces wind'' and ``Wind is used for \textit{producing electricity},'' which connect the question and answer.
QASC includes a gold, human-curated 2-hop explanation from the corpus for each question-answer pair.

Meanwhile, the Explainable QASC (eQASC) dataset adds 10 automatically generated explanations for each question-answer pair, each of which are labeled by annotators as valid or invalid \cite{jhamtani-clark-2020-learning}. 
While the state-of-the-art accuracy on QASC has reached up to 90\%,\footnote{See \url{https://allenai.org/data/qasc}.} only 76\% of questions have any valid explanation chains in eQASC. This indicates that \textit{explaining} the answers to questions in QASC is a more challenging problem than answering them. This motivates us to further explore the problem of generating multi-hop explanations for QASC.

\section{Methods}
In our experiments toward multi-hop explanation generation, we consider syntactic and semantic multi-hop retrieval methods, then explore ways to re-rank retrieved explanations to reduce the pool of candidates.

\subsection{Syntactic Methods}\label{sec:syntactic}
\begin{figure}
    \centering
    \includegraphics[width=0.4\textwidth]{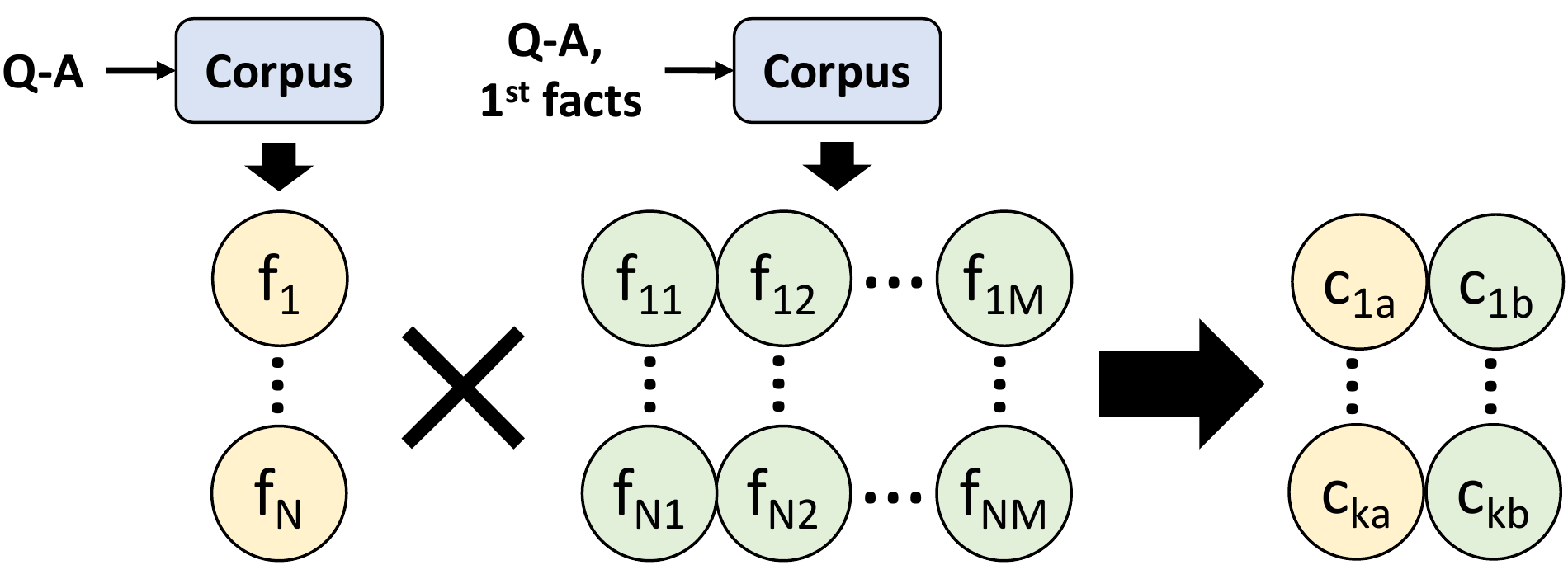}
    \vspace{-5pt}
    \caption{Syntactic pipeline used to generate multi-hop explanations in eQASC. First, the question-answer (Q-A) pair is used to query the ElasticSearch index for $N$ candidate first facts, each of which is used to query it for $M$ candidate second facts. All candidate first and second facts are paired, and the top-scored $K$ chains are returned as explanations. }
    \vspace{-1em}
    \label{fig:syntactic_pipeline}
\end{figure}

Syntactic information retrieval methods enable quick searching of millions of documents.
eQASC was originally generated using ElasticSearch, \footnote{https://www.elastic.co/} a fast but primarily syntactic search engine based on keyword overlap. After indexing the QASC corpus facts into an ElasticSearch index, \citet{jhamtani-clark-2020-learning} used a simple procedure (shown in Figure~\ref{fig:syntactic_pipeline}) to generate a 2-hop explanation for each question-answer pair from QASC. First, query the corpus for $N=20$ candidate first facts. For each candidate first fact, query the corpus for $M=4$ candidate second facts, where each candidate second fact must contain a word that appears in the question-answer pair, and a word that appears in the first fact. The purpose of this restriction is to force the resulting chain of facts to connect concepts in the question and answer through intermediate concepts. Lastly, from the set product of all candidate first facts and all candidate second facts, select up to $K=10$ candidate explanation chains, ranked by the sum of retrieval scores from the ElasticSearch engine.

\paragraph{Expanding syntactic retrieval.}
This is a simple, fast approach to generate a large number of candidate explanations. To improve the likelihood of generating a valid explanation, we can expand and diversify the search results by increasing $N$, $M$, and $K$. Specifically, we increase each of them to 200.

\subsection{Semantic Methods}\label{sec:semantic}
Alternatively, semantic information retrieval methods can enable stronger meaning representation than syntactic methods with a trade-off of search speed. Typical approaches generate a semantic vector embedding for all documents in a corpus. They then generate a comparable embedding of the query, and use vector similarity measures to rank documents.

\paragraph{Dense passage retrieval.}
Dense passage retrieval (DPR) is a recent approach to semantic information retrieval which learns dual encoders for queries and documents \cite{karpukhin-etal-2020-dense}. They are trained such that the query and document encoders generate similar embeddings for semantically similar queries and documents. Similarity is measured by inner product of vectors, and is maximized for matching queries and documents, but minimized for irrelevant queries and documents.
We can then use the document encoder to index the facts in a corpus, and efficiently query the index using the embedding from the query encoder.

For each question-answer pair in QASC and the two facts in its gold explanation chain, we can train a dense passage retriever to generate similar embeddings for the question-answer pair (query) and these facts (documents), then encode all facts in the QASC corpus for search purposes. 

\begin{figure*}
    \centering
    \includegraphics[width=0.7\textwidth]{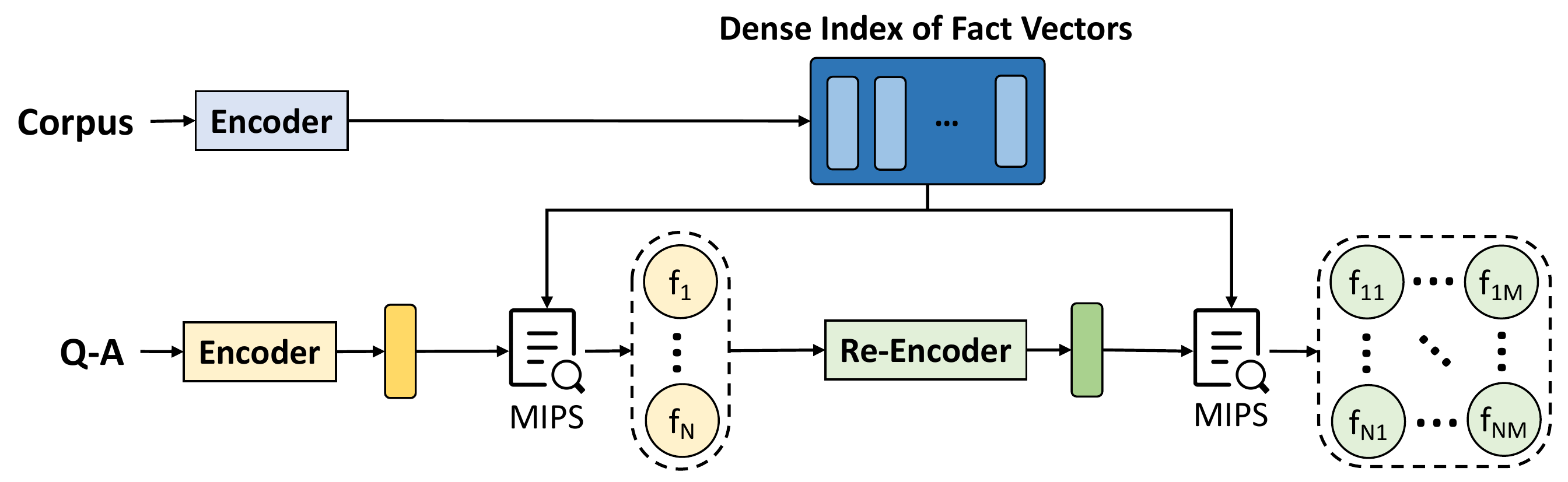}
    \vspace{-1em}
    \caption{Proposed semantic explanation pipeline. Facts are encoded using a fact encoder (in blue) and stored in a dense index, while the question-answer pair is encoded by a query encoder (in yellow). Maximum inner product search (MIPS) is used to query the index for $N$ candidate first facts, which are each re-encoded (in green), then used to query the index again for $M$ candidate second facts. All candidate first and second facts are paired, and the top-scored $K$ chains are returned as explanations. }
    \vspace{-1em}
    \label{fig:semantic expl pipeline}
\end{figure*}

\paragraph{Multi-hop reasoning.}
To generate a multi-hop explanation using this approach, we need to facilitate reasoning over the facts and queries. Given a question-answer pair from QASC, we first query the DPR index for {$N=5$} facts. To reduce error accumulation in generating the chain of facts, we then \textit{re-encode} each fact into a new query embedding incorporating the candidate fact and the original query. 

The re-encoder is a lightweight feedforward network inspired by a similar fact-translating function proposed in \citet{lin-etal-2021-differentiable}. Given the embeddings $q_{QA}$ and $d_1$ for the question-answer pair query and first fact document respectively, we use the gold explanation chains from QASC to learn the re-encoder $g(q_{QA}, d_1)$. Specifically, if $d_1$ is the embedding of the first fact in the gold explanation chain, we maximize the inner product between the re-encoded output $q_{r}$ and the document embedding $d_2$ for the second gold fact. Next, we query the DPR index again using $q_r$ to obtain {$M=2$} candidate second facts. To reduce noise, we filter out any facts that mention no concepts in either the question or answer.
Lastly, from the set product of all candidate first facts and all candidate second facts, select up to $K=10$ candidate explanation chains, ranked by the sum of retrieval scores, i.e., inner products when querying the DPR index. Our semantic explanation pipeline is shown in full in Figure~\ref{fig:semantic expl pipeline}. It is worth noting that our lightweight re-encoder operation can extend to any number of hops.

\subsection{Re-Ranking Candidate Explanations}
Both of our syntactic and semantic multi-hop retrieval systems can quickly propose candidate explanation chains for questions in QASC. However, both approaches over-generate candidates, and the high number of candidates (i.e., up to 200) limits the practical usefulness of the systems to an end user. As such, we lastly propose a re-ranker for candidates based on large-scale, pre-trained language models \cite{devlin-etal-2019-bert,liu2019roberta}. Specifically, we use the gold explanation chains from QASC to fine-tune a language model to the classification task of whether or not a candidate explanation is valid for a question-answer pair. We then re-rank a pool of candidates based on the system's estimated likelihood that each explanation chain is valid, and keep only the top $K=10$ candidates for a direct comparison to the syntactic approach used to generate eQASC.

\section{Experimental Results}
We next apply these approaches to the task of selecting 2-hop reasoning chains for question-answer pairs in QASC, and directly compare our results to the original procedure to generate eQASC. We compare systems by their individual \textit{gold retrieval rate} on the validation set for QASC, i.e., the percentage of question-answer pairs for which the gold explanation chain from QASC was successfully reproduced.\footnote{As the ordering of facts in QASC explanation chains does not typically matter, the gold retrieval rate counts both the forward and reverse forms of gold explanation chains.} This serves as an indicator of the quality of generated explanations, as it suggests that generated explanations tend to look more like those curated from the corpus by humans.

\subsection{Expanded Syntactic Explanation}
As mentioned earlier, we first expanded the ElasticSearch-based approach used to generate eQASC by increasing the search hyperparameters $N$, $M$, and $K$ each to 200. Selected results from this are listed in Table~\ref{tab:expand syntactic}. By only increasing $K$ (i.e., the number of candidate explanation chains considered) to 200, the retrieval rate increases from 31.1\% to 37.0\%. When increasing $N$ and $M$ (i.e., the number of candidate first and second facts considered) also to 200, the retrieval rate further increases to 46.5\%, a net 15.4\% gain.

\begin{table}
    \centering
    \footnotesize
    \begin{tabular}{ccc|c}
        \toprule
        \textbf{N} & \textbf{M} & \textbf{K} & \textbf{Gold Retrieval Rate (\%)} \\\midrule
        20 & 4 & 10 & 31.1 \\
        20 & 4 & 200 & 37.0 \\
        200 & 200 & 200 & \textbf{46.5} \\
        \bottomrule
    \end{tabular}
    \normalsize
    \caption{Gold explanation chain retrieval rates for syntactic multi-hop retrieval with ElasticSearch on QASC validation set. The first row indicates the original search hyperparameters used to generate eQASC, while the last two rows increase hyperparameters to expand and diversify the search.}
    \vspace{-1em}
    \label{tab:expand syntactic}
\end{table}

\subsection{Syntactic-Semantic Multi-Hop Explanation}
Next, we incorporate our semantic multi-hop retrieval process powered by DPR. 

\paragraph{Training details.}
The dual encoders for DPR are learned starting from pre-trained \textsc{BERT}-base~\cite{devlin-etal-2019-bert}. The best encoders are selected based on the mean squared error between embeddings for matching question-answer pairs and facts on the QASC validation set. The batch size is fixed at 16, while learning rate and number of training epochs are selected based on a grid search. 
For the re-encoder, the training batch size, learning rate, and number of epochs are similarly selected based on a grid search, minimizing the mean squared error between the output re-encoded queries and target fact embeddings on the validation set. 

\paragraph{Results.}
Table~\ref{tab:semantic} compares the gold retrieval rate of various combinations of the syntactic and semantic approaches for multi-hop retrieval on the QASC validation and testing sets.\footnote{When combining the syntactic and semantic approaches, we replace up to the lowest-ranked 25\% of syntactic candidate explanation chains with the top semantic candidate explanation chains.} Our results show that while using only the semantic candidate explanation chains leads to a 13.9\% gold retrieval rate at best, combining the expanded syntactic and semantic candidates gives us the best result of up to 51.1\% gold retrieval rate, outperforming the case where only the expanded syntactic candidates are considered. Thus, the semantic approach finds some of the missing gold explanations that the syntactic approach misses, suggesting that both syntactic and semantic approaches are needed for generating the best-quality explanations on QASC questions.

\begin{table}
    \centering
    \footnotesize
    \begin{tabular}{c|cc}
    \toprule
        \textbf{Approach}  & \multicolumn{2}{c}{\textbf{Gold Retrieval Rate (\%)}} \\
         & \textit{Validation} & \textit{Test} \\\midrule
        syntactic & 37.0 & 40.2 \\
        syntactic (exp.) & 46.5 & 49.3 \\
        semantic & 10.8 & 13.9 \\
        syntactic (exp.) + semantic & \textbf{49.9} & \textbf{51.1} \\
        \bottomrule
    \end{tabular}
    \normalsize
    \caption{Gold explanation chain retrieval rates (top $K=200$ candidates) for combinations of multi-hop retrieval approaches on QASC. Syntactic refers to the second result from Table~\ref{tab:expand syntactic}, while syntactic (exp.) refers to the expanded third result. Semantic refers to the previously introduced DPR-based multi-hop retrieval approach.}
    \vspace{-1em}
    \label{tab:semantic}
\end{table}

\subsection{LM Re-Ranking}
While our results improve the gold retrieval rate by a wide margin, recall that our multi-hop retrieval approaches for QASC increase the number of candidate explanation chains $K$ to 200. Such a large set of candidates is not useful in practice, as a human user would have to sort through a cumbersome number of explanations in order to judge the machine's understanding of the question and answer. As such, we lastly present our experiments on \textit{re-ranking} candidate explanation chains, which enables us to truncate our results to $K=10$ top candidate explanation chains without massive performance drops, and consequently compare our approach directly to the original approach used to generate eQASC.

\begin{table}
    \centering
    \footnotesize
    \begin{tabular}{c|c|cc}
    \toprule
        \textbf{Retrieval Approach}  & \textbf{Re-Ranker} &\multicolumn{2}{c}{\textbf{Gold RR (\%)}} \\
         & & \textit{Val.} & \textit{Test} \\\midrule
        syntactic & -- & 31.1 & 34.1 \\\midrule
        syntactic (exp.) & \textsc{BERT} & 36.3 & 34.0 \\
        syntactic (exp.) + semantic & \textsc{BERT} & {36.4} & {34.1} \\\midrule
        syntactic (exp.) & \textsc{RoBERTa} & 37.9 & 36.2 \\
        syntactic (exp.) + semantic & \textsc{RoBERTa} & \textbf{38.1} & \textbf{36.4} \\        
        \bottomrule
    \end{tabular}
    \normalsize
    \caption{Gold explanation chain retrieval rates (RR; top $K=10$ candidates) for combinations of multi-hop retrieval approaches on QASC, re-ranked by fine-tuned language models. Syntactic refers to the original approach used to generate eQASC, while syntactic (exp.) refers to the expanded third result from Table~\ref{tab:expand syntactic}. Semantic refers to our proposed DPR-based multi-hop retrieval approach.}
    \vspace{-1em}
    \label{tab:reranking}
\end{table}

\paragraph{Training details.}
Using the re-ranking approach described earlier, we fine-tune the \textsc{BERT}~\cite{devlin-etal-2019-bert} and \textsc{RoBERTa}~\cite{liu2019roberta} 
pre-trained language models.\footnote{For both models, we use the ``base'' form which has 12 hidden layers, a hidden dimension of 768, and 12 attention heads.} Models are trained with a 1:2 ratio of gold and invalid explanation chains, with 3 unique invalid explanation chains randomly sampled from ElasticSearch results per gold explanation chain (forward and reverse forms). Models are selected based on instances achieving the highest top-1 gold retrieval rate, i.e., proportion of question-answer pairs where the gold explanation chain is ranked highest, on the QASC validation set similarly redistributed in this way. 


\paragraph{Results discussion.}
Table~\ref{tab:reranking} compares the final gold retrieval rates for the top $K=10$ re-ranked candidates from various approaches. While the original syntactic approach for generating eQASC achieves a respective 31.1\% and 34.1\% 
gold retrieval rate on the validation and testing sets, our expanded syntactic approach achieves up to 37.9\% and 36.2\% gold retrieval rate with \textsc{RoBERTa}. Again with \textsc{RoBERTa}, our syntactic-semantic multi-hop retrieval achieves the best results of 38.1\% and 36.4\% on the validation and testing sets, respectively, exceeding the baselines. 
After narrowing down from 200 candidate explanations to 10 with the re-ranker, we retain up to a 7.0\% net improvement of gold retrieval rate compared with the baseline. Given that the net gain was 13.9\% with 200 candidate explanations, one future direction is to improve the re-ranker performance, so that we can retain more of this improvement.
To achieve this, one option is to revisit the re-ranker training, which did not incorporate negative examples proposed by DPR, and may experience generalization issues.

\section{Conclusion}
In this work, by utilizing a small amount of ground truth supervision, we explored approaches to improve the generation of multi-hop explanations from a corpus of declarative facts. We show that both fast, syntactic methods and slow, semantic methods are useful for gathering relevant evidence for explanation. 
To facilitate multi-hop reasoning from one piece of evidence to the next, we had some success in using a lightweight feedforward re-encoder, as opposed to state-of-the-art graph-based approaches that consume too much time and memory for practical online use.
As many of our approaches over-generate candidate explanations, we lastly explored using pre-trained language models to re-rank and filter candidates. Our results suggest this is a significant challenge, and future work may further explore this problem.

\bibliography{main}

\end{document}